\documentclass[conference]{IEEEtran}
\IEEEoverridecommandlockouts
\usepackage{cite}
\usepackage{amsmath,amssymb,amsfonts}
\usepackage{algorithmic}
\usepackage{graphicx}
\usepackage{textcomp}
\usepackage{xcolor}
\def\BibTeX{{\rm B\kern-.05em{\sc i\kern-.025em b}\kern-.08em
    T\kern-.1667em\lower.7ex\hbox{E}\kern-.125emX}}
\begin{document}

\title{Learning Multi-Access Point Coordination \\in Agentic AI Wi-Fi with Large Language Models
}

\author
{\IEEEauthorblockN
  {Yifan Fan\textsuperscript{1}, Le Liang\textsuperscript{1,2}, Peng Liu\textsuperscript{3}, Xiao Li\textsuperscript{1}, Ziyang Guo\textsuperscript{3}, Qiao Lan\textsuperscript{3}, Shi Jin\textsuperscript{1}, and Wen Tong\textsuperscript{3}
\IEEEauthorblockA{\textrm{\textsuperscript{1}School of Information Science and Engineering, Southeast University, Nanjing 210096, China} \\
\textrm{\textsuperscript{2}Purple Mountain Laboratories, Nanjing 211111, China} \\
\textrm{\textsuperscript{3}Wireless Technology Lab, 2012 Labs, Huawei Technologies Co., Ltd,} \\
E-mail: $\left\{\textrm{yifan\_fan, lliang, li\_xiao, jinshi}\right\}$@seu.edu.cn;}
  }
}

\maketitle

\begin{abstract}
Multi-access point coordination (MAPC) is a key technology for enhancing throughput in next-generation Wi-Fi within dense overlapping basic service sets. However, existing MAPC protocols rely on static, protocol-defined rules, which limits their ability to adapt to dynamic network conditions such as varying interference levels and topologies. To address this limitation, we propose a novel Agentic AI Wi-Fi framework where each access point, modeled as an autonomous large language model agent, collaboratively reasons about the network state and negotiates adaptive coordination strategies in real time. This dynamic collaboration is achieved through a cognitive workflow that enables the agents to engage in natural language dialogue, leveraging integrated memory, reflection, and tool use to ground their decisions in past experience and environmental feedback. Comprehensive simulation results demonstrate that our agentic framework successfully learns to adapt to diverse and dynamic network environments, significantly outperforming the state-of-the-art spatial reuse baseline and validating its potential as a robust and intelligent solution for future wireless networks.
\end{abstract}


\section{Introduction}

The upcoming IEEE 802.11bn standard, or Wi-Fi 8, introduces multi-access point coordination (MAPC) as a key mechanism to enhance performance in dense Wi-Fi deployments \cite{galati2024primer}. Specifically, MAPC enables neighboring access points (APs) in overlapping basic service sets (OBSS) to jointly manage radio resources, thereby mitigating the adverse impact of co-channel interference and boosting network throughput. This is achieved through various coordination paradigms, such as coordinated time division multiple access (Co‑TDMA) and coordinated spatial reuse (Co‑SR) \cite{tgbn}. These schemes are typically realized via a trigger-frame-based mechanism, where a sharing AP orchestrates a coordinated transmission opportunity (TXOP) by broadcasting control information to shared APs \cite{cosr}. This mechanism has proven effective, with studies showing that trigger-enabled Co-SR can boost aggregate throughput by up to 59\% compared to legacy Wi-Fi protocols \cite{Wilhelmi2023thr}.

However, the static and protocol-defined nature of conventional MAPC limits its ability to adapt to dynamic conditions, since it is confined to a set of pre-defined, reactive heuristics. These constraints have motivated the development of artificial intelligence (AI)-based approaches that enable APs to learn flexible and efficient scheduling strategies by discovering patterns from experience, rather than merely executing fixed-protocol designs. In \cite{wojnar2024ieee}, the hierarchical multi-armed bandits algorithm is employed to enhance Co-SR scheduling by learning effective AP-station (STA) grouping strategy for concurrent transmissions, yielding an approximate 80\% gain in aggregate throughput over legacy IEEE 802.11. Similarly, a hierarchical multi-agent reinforcement learning (HMARL)-based MAPC algorithm is developed in \cite{yu2025hmarl}, where a high-level policy selects STAs to transmit and a low-level policy controls transmit power to enable efficient Co-SR scheduling. Simulation results show that HMARL outperforms conventional schemes in terms of throughput and latency, while also maintaining robustness when coexisting with legacy APs. 

However, existing AI-driven MAPC approaches are primarily built on agents powered by small, specialized AI models. These task-specific agents typically rely on simple reward-driven decision logic and are limited to narrow coordination modes. Furthermore, these approaches often exhibit poor generalization to unseen environments, as their effectiveness heavily relies on specific training scenarios. The advent of large language models (LLMs) and the subsequent rise of agentic AI offer a promising avenue to overcome these limitations. Unlike smaller, task-specific models, LLM agents leverage extensive world knowledge and advanced reasoning to achieve complex goals. The potential of LLM has already been demonstrated in wireless communications \cite{liang2025wirelessllm}. For instance, single-agent frameworks like WirelessAgent \cite{tong2025wirelessagent} have proven that LLM can apply its inherent reasoning capabilities to interpret complex network states and devise adaptive network configurations.

While a single-agent approach is promising for centralized control, MAPC is inherently a distributed control problem requiring collaboration among multiple APs. In response, we propose a novel MAPC protocol based on a collaborative multi-LLM-agent system. Our framework is built upon multi-agent natural language dialogue, through which each agent leverages its integrated reasoning, memory, and reflection capabilities to derive optimal coordination schemes for diverse and dynamic interference scenarios. Our simulation results demonstrate superior adaptability and coordination efficiency, achieved through self-organized agent negotiation rather than rigid heuristics, thus establishing multi-LLM-agent systems as a promising frontier for developing truly intelligent and scalable wireless communication.

\section{System Model}

\begin{figure*}[htbp]
\centerline{\includegraphics[width=\linewidth]{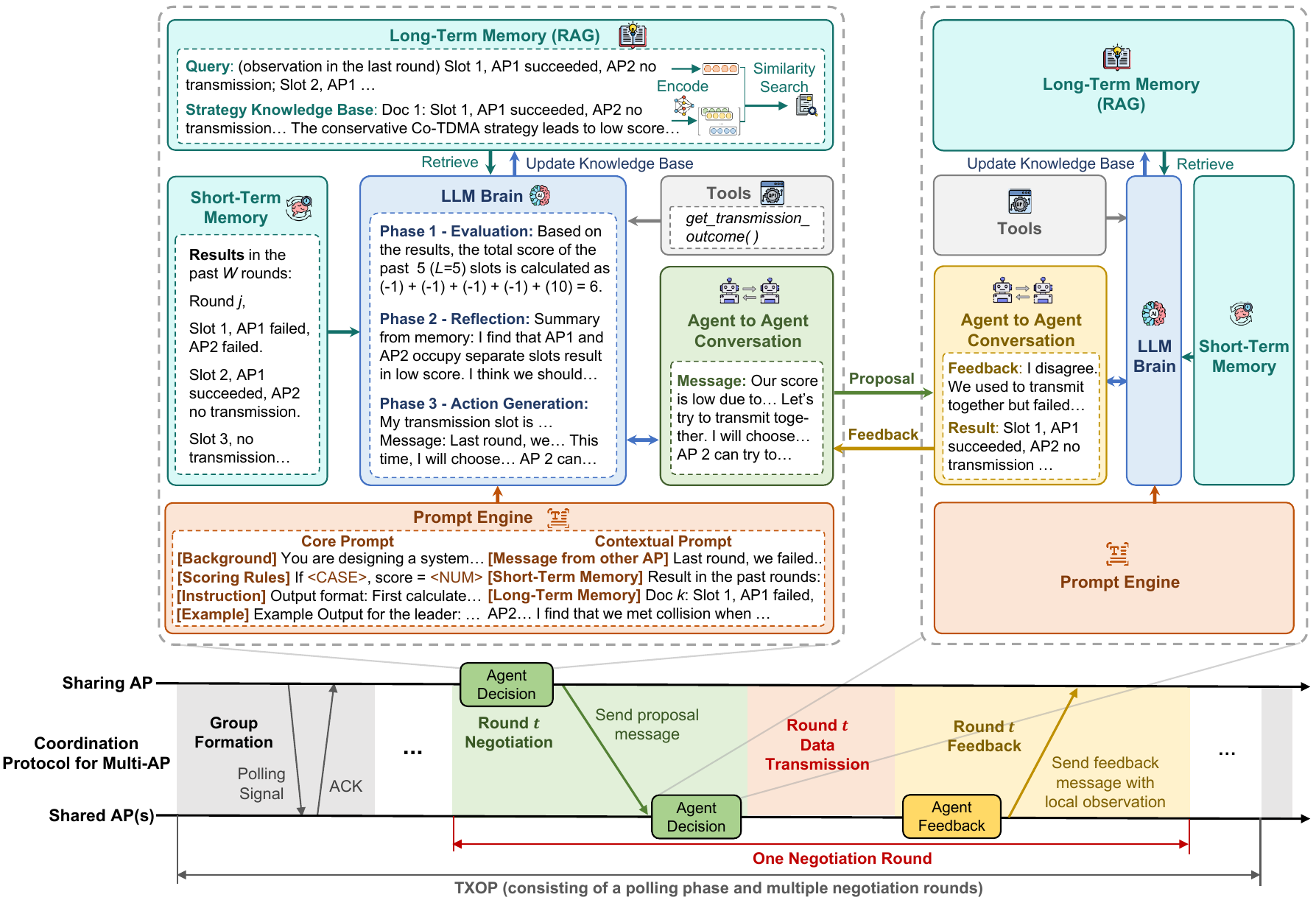}}
\caption{Agentic workflow and coordination protocol for multi-AP cooperation.}
\label{protocol}
\end{figure*}

We consider the downlink communication of a Wi-Fi OBSS system comprising $K$ basic service sets (BSSs), where each AP serves one associated STA. Building upon the distributed MAPC framework, our system model defines a structured coordination process that unfolds in a series of rounds, establishing the operational rules for our agentic algorithm. The overall workflow of this proposed protocol is illustrated in Fig.~\ref{protocol}. As depicted in the bottom panel of Fig.~\ref{protocol}, the coordination process is initiated when an AP wins channel access via the standard carrier-sense multiple access with collision avoidance (CSMA/CA) procedure and secures a TXOP. The coordination activities then unfold within this secured TXOP. First, the winning AP, now the designated sharing AP, initiates a polling phase to establish a coordination group with a set of neighboring shared APs. Following this group formation, the remainder of the TXOP is structured to contain $T$ consecutive negotiation rounds, allowing for a sustained period of multi-round coordination without repeated channel contention and group formation. Each negotiation round is composed of $L$ discrete time slots, with each slot's duration corresponding to a single packet transmission. Within each negotiation round, each AP acts as an agent, and its fundamental decision is to select a transmission schedule, formalized by a binary vector $\mathbf{a}_k = [a_{k,1}, a_{k,2}, \dots, a_{k,L}] \in \{0, 1\}^L$, where $a_{k,\ell}=1$ signifies that AP $k$ transmits in the $\ell$-th time slot, and $a_{k,\ell}=0$ indicates no transmission. To coordinate their actions, the agents engage in a two-way dialogue within each round. The process begins with the sharing AP broadcasting a coordination message to guide the decision-making of the shared APs. After the round concludes, the shared APs report their outcomes back to the sharing AP, completing the feedback loop. The specific design of our agentic algorithm, based on this MAPC protocol model, is detailed in the following section.

The dense deployment of APs creates co-channel interference, which can significantly degrade signal quality and lead to transmission failures. This situation gives rise to a core coordination problem: How to balance two competing objectives: interference avoidance, best achieved via Co-TDMA, and throughput maximization, the primary goal of Co-SR. These two strategies are formally characterized as follows:

\begin{itemize}
    \item \textbf{Co-TDMA}: A conservative, interference-free strategy achieved when agents choose non-overlapping time slots, i.e., $\sum_k a_{k,\ell} \leq 1$ for all $\ell$. This guarantees successful transmissions but limits spectral efficiency.
    \item \textbf{Co-SR}: An aggressive strategy that boosts throughput by allowing concurrent transmissions, i.e., $\sum_k a_{k,\ell} > 1$ for some $\ell$. However, it risks performance degradation or outright failure if interference is too high.
\end{itemize}

Our goal is to develop a multi-LLM-agent system that can learn an adaptive coordination policy, guiding the agents to autonomously navigate the Co-SR/Co-TDMA trade-off by reasoning about the environment through interaction, ultimately aiming to maximize the overall network throughput.

\section{Agentic Workflow for Multi-AP Coordination}

In this section, we detail the design of our proposed agentic algorithm, which operates within the proposed MAPC framework. Our approach empowers each AP with a sophisticated LLM agent, enabling them to surpass fixed protocols and engage in intelligent, adaptive coordination. We first detail the multi-agent coordination protocol that governs their interaction, then delve into the cognitive architecture that enables each agent's intelligent participation, and finally analyze the emergent coordination behaviors that arise from their synergy.

\subsection{The Multi-Agent Coordination Protocol}

The core of our algorithm is a multi-agent natural language dialogue that unfolds over negotiation rounds within a TXOP. The process for each round is illustrated in the bottom panel of Fig.~\ref{protocol}, and proceeds as follows.

\textbf{1) Group Formation:} At the beginning of the TXOP, the sharing AP sends a polling signal to its neighboring APs, which respond with an acknowledgment (ACK) signal to confirm cooperation.

\textbf{2) Proposal Generation by the Sharing AP:} In each round, the sharing AP's agent leverages its reasoning capabilities to analyze the coordination history and perceived interference patterns. Based on this analysis, it first determines its own transmission schedule and then generates a natural language message proposing a collaborative strategy for the group.

\textbf{3) Proposal Evaluation by the Shared APs:} Upon receiving the message, the shared AP does not merely comply. Instead, it combines the proposal with its own local knowledge and autonomously determines its transmission schedule.

\textbf{4) Data Transmission:} All APs execute their chosen transmission schedules for the $L$ slots of the current round. Afterwards, the agents invoke external tools to perceive the transmission results.

\textbf{5) Feedback and Strategy Refinement:} The shared APs then report their individual outcomes back to the sharing AP, providing it with a complete picture of the round's performance. This collective feedback triggers a reflection process in each agent, allowing them to refine their strategies for the subsequent negotiation rounds, creating a continuous cycle of learning and improvement within the TXOP.

\subsection{LLM Agent Cognitive Architecture}

To realize this intelligent workflow, each AP is modeled as an autonomous LLM agent with a modular internal architecture, as depicted in the upper part of Fig.~\ref{protocol}. At its core lies the LLM brain, supported by memory modules, environmental interaction tools, and a prompt engine, which together enable the agent to perceive, reason, and act adaptively within the multi-AP environment.

\subsubsection{The Brain and its Reasoning Workflow}
The brain of the agent is a pre-trained LLM that leverages its reasoning capabilities to interpret the complex, dynamic state of the multi-AP environment and devise effective coordination strategies based on the coordination message and the historical transmission outcomes. To navigate the complex MAPC problem, the agent employs advanced prompting techniques such as in-context learning (ICL) \cite{icl} and chain-of-thought (CoT) \cite{cot}. ICL enables the agent to learn from examples provided directly within the prompt, allowing it to quickly adapt to specific environmental interference patterns by recognizing recurring situations without costly fine-tuning. CoT prompting, on the other hand, enhances the agent's logical reasoning capabilities by guiding it to break down the complex coordination task into a series of structured reasoning steps, including evaluation, reflection, and action generation:

\textbf{Evaluation}: The agent first performs an evaluation guided by a performance scoring mechanism defined within the prompt. The evaluator LLM analyzes feedback from the previous round, such as collision reports of all APs, to calculate a performance score. This score is designed to holistically quantify the outcome of the last round, balancing factors like achieved throughput against the negative impact of collisions. For example, successful simultaneous transmissions receive a positive reward, collisions are penalized, and idle slots incur a penalty to discourage spectrum underutilization.

\textbf{Reflection}: Following the evaluation, the agent enters the reflection phase, central to both immediate decision-making and long-term self-improvement. This phase operates as an inner monologue where the agent performs CoT reasoning to synthesize the recent transmission outcomes with its past experiences, thereby formulating a high-level strategy for the current situation. The agent then curates its long-term memory by exclusively storing the strategies from high-performing rounds, enabling a continuous cycle of self-improvement.

\textbf{Action Generation}: Finally, the LLM agent translates the strategy into a concrete, actionable output. It generates the final text, which includes its own transmission schedule and the natural language message to be broadcast to other agents.

\subsubsection{Memory Module}
To enable learning and adaptation over time, each agent is equipped with a hybrid memory system that supports both immediate tactical adjustments and long-term strategic evolution.

The short-term memory maintains a sliding window of the most recent $W$ negotiation rounds, storing a concise history of the agents' actions and their corresponding transmission outcomes. This recent history is directly injected into the LLM's prompt at the beginning of each decision cycle. This mechanism serves two critical purposes. First, it prevents the agent from repeatedly making the same mistakes in consecutive rounds. Second, it allows the agent to perceive the short-term dynamics of the wireless environment. Since the network topology remains relatively stable over short periods, this memory window helps the agent distinguish between an anomalous, one-off transmission failure and a persistent interference pattern. For instance, a single collision might not trigger a drastic strategic shift, but observing collisions across several recent rounds with the same set of APs will strongly signal the need for a more conservative approach. This stabilizes the agent's behavior, making it robust against random fluctuations.

The long-term memory functions as a persistent, evolving knowledge base, implemented via a retrieval-augmented generation (RAG) method. It stores a curated collection of best-practice exemplars, where each exemplar consists of a historical situation description, the high-performing action, the performance score, and the corresponding strategic reflection that justifies the decision. Specifically, the agent uses the current situation as a query to retrieve the most similar historical exemplars from this knowledge base, with relevance quantified by cosine similarity computed over embedding representations. These retrieved examples are then formatted as few-shot demonstrations and inserted into the prompt. In doing so, the LLM is inspired to generate more sophisticated and contextually appropriate strategies.

To enable continuous learning and strategic growth, the knowledge base evolves through a contextualized performance-based update protocol. Initially, the knowledge base contains a set of hand-crafted exemplars representing good practices covering several typical coordination scenarios. After each round, the agent first retrieves a cluster of the most similar historical exemplars from the current knowledge base. If no similar precedents are found with similarity scores above a pre-defined threshold, the new experience is automatically added as it represents an unexplored scenario. Otherwise, the performance score of the current round is compared against the lowest score within that cluster. This new experience is retained only if it demonstrates superior performance, potentially replacing the cluster's lowest-scoring member if the memory is at maximum capacity. This update method ensures the knowledge base is continuously curated with the best-known strategies for a diverse range of coordination scenarios.

\subsubsection{Tool Use Module}
Tool use is a critical mechanism that enables the agent to perceive and interact with the external environment through standardized interfaces. For instance, at the end of the data transmission phase of each negotiation round, the agent invokes a \textit{get\_transmission\_outcome()} tool. This function interfaces with the network monitoring system to retrieve feedback and translates this raw data into a structured text string, allowing LLM agents to read, understand, and reason about the real-world consequences of its actions.

\subsubsection{Prompt Design}

The LLM brain is powered by a meticulously designed two-part prompt. The first part, a static core prompt, defines the agent's long-term identity and operational rules, including background, performance scoring rules, format instructions, and ICL examples. The second part, a dynamic contextual prompt, is reconstructed each round to provide contextual information by incorporating real-time inputs such as messages from other APs, short-term memory, and relevant exemplars retrieved from long-term memory.

\subsection{Emergent Coordination Behaviors}

The agentic architecture described above facilitates a new paradigm of dynamic coordination. Unlike traditional systems with fixed rules, our agents collaboratively determine the optimal coordination strategy through dialogue. This coordination is manifested in two key aspects:

\textbf{Dynamic Switching of Coordination Strategies}: The primary form of high-level coordination is the adaptive selection between Co-TDMA and Co-SR, which is determined dynamically by the agents rather than being pre-programmed. Through the continuous cycle of action, feedback, and reflection, the agents learn to recognize the underlying interference patterns of their network environment. For example, a history of collisions guides the agent to adopt a conservative CoTDMA strategy, while a record of successful transmissions encourages a more aggressive Co-SR approach. This ability to flexibly switch between conservative and aggressive coordination modes, based on learned experience rather than fixed rules, is a core strength of our agentic approach.

\textbf{Negotiation Protocol in Natural Language}: Beyond selecting a high-level coordination scheme, the agentic dialogue facilitates fine-grained negotiation and optimization within that chosen scheme. Unlike simple control signals, the natural language messages are rich, contextual proposals that encapsulate strategic intent. For example, instead of just broadcasting a fixed schedule, a sharing AP can formulate a proposal that includes not only its own intended action but also contingent suggestions or queries for its peers. A shared AP, upon receiving such a message, does not passively execute a command. Instead, it engages its own reasoning module to evaluate the proposal against its local observations and knowledge. It can then autonomously decide to accept the suggestion, reject it, or even formulate a counter-proposal, enabling a level of dynamic, nuanced collaboration that is unattainable with rigid, pre-defined protocols.

\section{Simulation Results}

\subsection{Simulation Setup}

Our simulations are conducted in a Wi-Fi OBSS downlink transmission scenario comprising $K$ BSSs, with a single STA associated with each AP. All APs are assumed to have saturated downlink traffic. We employ a standard channel model to capture propagation effects. Large-scale fading is modeled using the IEEE 802.11ax residential path loss model \cite{residential} combined with log-normal shadowing, while small-scale fading follows a Nakagami-$m$ distribution with $m=1.5$. We assume that the small-scale fading remains constant for the duration of a single packet transmission.

To evaluate the long-term strategic learning of our agents, each simulation run consists of $T$ consecutive negotiation rounds. Each round has a duration of 400 $\mu s$ for data transmission, which is discretized into $L=5$ equal-length time slots. Within each round, the agents autonomously decide which of these $L$ slots to utilize for transmission.

To create a diverse set of network topologies, the APs are randomly distributed within a defined area, and the associated STA for each AP is located at a random distance and a random orientation from it. For analytical purposes, we categorize these specific topologies into two representative scenarios.

\begin{itemize}
    \item \textbf{Co-TDMA-Favored Scenario}: In this setup, APs are positioned in close proximity, resulting in high inter-AP interference where Co-TDMA is the more viable strategy.
    \item \textbf{Co-SR-Favored Scenario}: In this setup, APs are placed far apart, leading to low interference levels that permit and encourage the use of Co-SR strategy.
\end{itemize}

To demonstrate the general applicability of our proposed framework, we conduct experiments using two distinct LLMs to power the agents: GPT-4o \cite{openai2024gpt4ocard} and DeepSeek-R1 \cite{deepseekai2025deepseekr1}. The multi-agent dialogue and workflow are implemented using AutoGen \cite{wu2023autogen}, a framework for multi-agent conversational applications developed by Microsoft. For the agent's cognitive architecture, the short-term memory is configured with a sliding window of $W=5$ rounds, and the long-term memory's knowledge base is set to a maximum capacity of 10 exemplars.

\subsection{Simulation Results}

\begin{figure}[htbp]
\centerline{\includegraphics[width=1\linewidth]{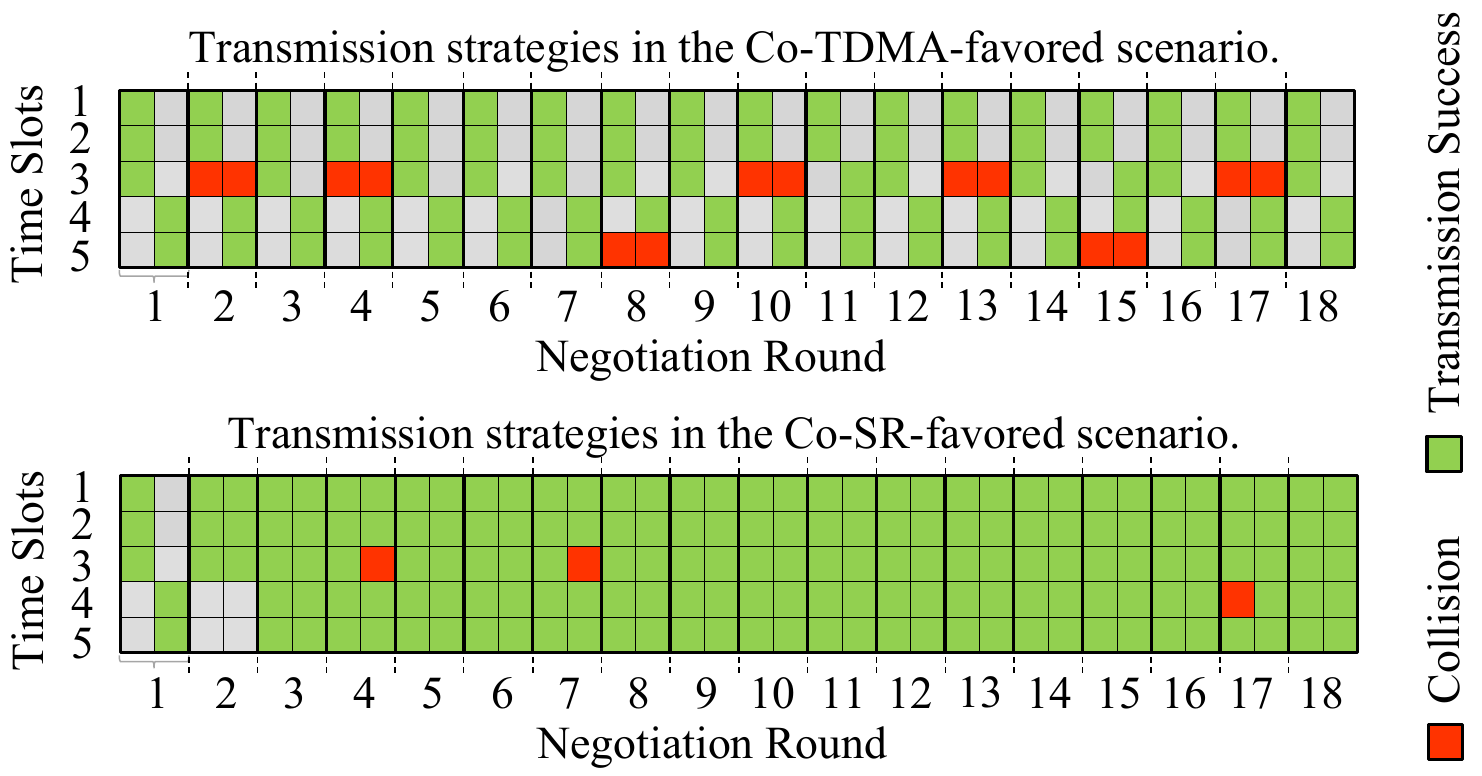}}
\caption{Transmission strategies of the agentic MAPC protocol under the Co-TDMA-favored and Co-SR-favored scenarios. For each negotiation round, the two columns within the bold frame display the transmission states in the 5 slots of the sharing AP (left) and the shared AP (right), respectively.}
\label{2ap-case}
\end{figure}

We first analyze the agents' behavior in the 2-AP scenarios. Fig.~\ref{2ap-case} illustrates the slot-by-slot transmission schedules adopted by the DeepSeek-R1-powered agents over 18 rounds, across both Co-TDMA-favored and Co-SR-favored topologies. As shown in Fig.~\ref{2ap-case}, the agents begin with a conservative Co-TDMA strategy to safely probe the environment. From this starting point, their collaborative strategy then evolves differently based on the specific interference conditions. In the high-interference scenario, the agents' strategy evolves into a hybrid schedule that balances the risk of a single-slot collision with the potential throughput gains from Co-SR. This explorational behavior strongly suggest the potential for adapting to dynamic changes in channel interference. Conversely, in the low-interference scenario, the agents progressively explore more aggressive options until they converge on a full Co-SR schedule to maximize throughput. The results reveal that the agent can develop sophisticated and interference-aware coordination strategies. Moreover, we have confirmed that effective coordination is also achievable in a heterogeneous setting where each AP is powered by a different LLM, such as GPT-4o and DeepSeek-R1, highlighting the framework's robustness and potential for interoperability in multi-vendor environments.

\begin{table}[htbp]
\caption{Comparison of normalized throughput between Wi-Fi 6 spatial reuse and agentic protocol across various topologies.}
\begin{center}
\begin{tabular}{|c|c|c|c|c|c|}
\hline
\textbf{} & \textbf{AC\_BE} & \textbf{GPT-4o} & \textbf{DeepSeek} \\
\hline
\textbf{2AP Co-TDMA-Favored} & 0.84 & \textbf{1.00} & 0.91 \\
\hline
\textbf{2AP Co-SR-Favored} & 1.05 & 1.85 & \textbf{1.87} \\
\hline
\textbf{3AP Co-TDMA-Favored} & 0.86 & \textbf{0.91} & 0.90 \\
\hline
\textbf{3AP Co-SR-Favored} & 1.35 & \textbf{2.06} & 1.80 \\
\hline
\end{tabular}
\label{wifi6compare}
\end{center}
\end{table}

To further validate our framework's effectiveness, we compare its throughput performance against a conventional \mbox{Wi-Fi 6} OBSS Packet-Detect (OBSS/PD) spatial reuse scheme, with the traffic of best-effort access category (AC\_BE). For the Wi-Fi~6 baseline, we establish a robust benchmark by performing an exhaustive search over various OBSS/PD sensitivity threshold and transmit power settings, selecting the optimal parameter combination for each scenario. All entries in Table~\ref{wifi6compare} are normalized throughput values, defined as the total proportion of time used for successful transmissions across all APs in the network. The result clearly shows that our agentic approach outperforms the optimized baseline in both the Co-TDMA-favored and Co-SR-favored scenarios, with a particularly pronounced advantage where the spatial reuse potential is high. This superiority stems from the inherent flexibility of our agentic approach. While the conventional method is confined to a discrete set of pre-defined parameters, our agents can discover more granular and effective hybrid slot-by-slot coordination strategies, thus learning more expressive and efficient collaborative behaviors.

\begin{figure}[htbp]
\centerline{\includegraphics[width=0.92\linewidth]{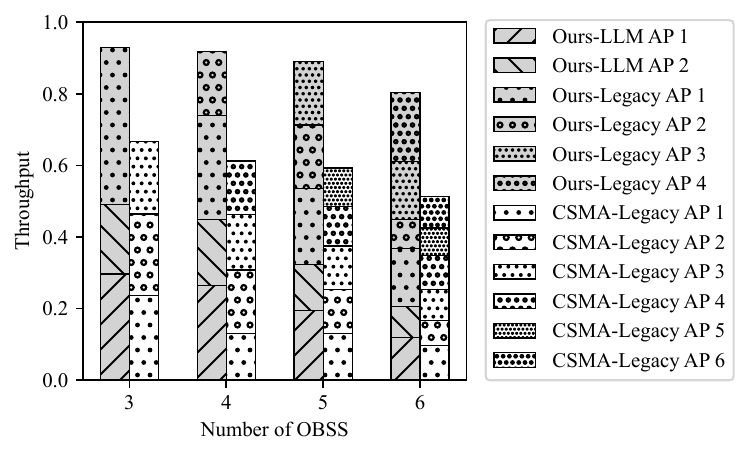}}
\caption{Coexistence between LLM-driven APs and legacy CSMA/CA APs in the Co-TDMA-favored scenario.}
\label{legacy}
\end{figure}

We also assess the framework's ability to coexist with legacy systems. We simulate a mixed-network scenario where two of our agentic APs operate alongside several legacy Wi-Fi APs. In this setup, all nodes contend for the channel via the standard CSMA/CA mechanism. The results in Fig.~\ref{legacy} demonstrate that the introduction of our agentic framework does not disrupt the normal packet transmission of the legacy Wi-Fi APs, highlighting its strong backward compatibility and potential for seamless integration into existing networks. Moreover, we also compare the throughput performance of our approach against a conventional CSMA/CA baseline, which is represented by the legacy APs configured with a clear channel assessment (CCA) threshold of -82 dBm and a transmit power of 20 dBm. The comparison reveals that our agentic framework achieves higher throughput even in a mixed environment by eliminating the back-off delay inherent in CSMA/CA schemes.

\begin{table}[htbp]
\caption{Normalized throughput performance in ablation study across different scenarios.}
\begin{center}
\begin{tabular}{|c|c|c|}
\hline
\textbf{Method} & \textbf{2AP Co-TDMA} & \textbf{2AP Co-SR} \\ \hline
Ours & \textbf{1.00} & \textbf{1.85} \\ \hline
w/o Reflection Module & 1.00 & 1.00 \\ \hline
w/o Inter-Agent Negotiation & 0.4 & 1.70 \\ \hline
w/o Short-Term Memory & 0.71 & 1.81 \\ \hline
w/o Long-Term Memory & 0.66 & 1.75 \\ \hline
\end{tabular}
\label{ablation}
\end{center}
\end{table}

Finally, we conduct an ablation study to investigate the contribution of each key component in our architecture by systematically disabling individual modules, with the corresponding normalized throughput presented in Table~\ref{ablation}. This study reveals the distinct role that each module plays in enabling effective coordination among agents. First, disabling the reflection module severely curtails exploration, as the agent lacks the ability to abstract from raw transmission outcomes and to reason about long-term strategies. Without this deeper reasoning capability, after achieving an initial success with Co-TDMA, the agent quickly converges to this safe baseline and ceases to explore higher-scoring Co-SR strategies. Second, when inter-agent negotiation is removed, each agent becomes isolated and can only rely on its own local performance history. This leads to selfish and myopic decision-making, where agents pursue individual gains without regard to global coordination. The lack of shared information prevents them from aligning their transmission schedules, which in turn results in suboptimal throughput despite the presence of advanced reasoning mechanisms. Third, the absence of short-term memory introduces another vulnerability. In particular, under high-interference scenarios, the agents may repeatedly attempt Co-SR despite consecutive failures, as they immediately forget the negative feedback from prior collisions. This short-sightedness destabilizes their behavior and reduces overall performance. Finally, disabling the RAG-based long-term memory highlights its importance in enhancing strategic reasoning, as the RAG mechanism augments the LLM's logic by providing contextually relevant, high-performing exemplars. This module is particularly effective in handling more complex scenarios. Taken together, these results clearly validate the necessity of each component for achieving robust and intelligent coordination.

\section{Conclusion}

In this paper, we introduced a novel Agentic AI Wi-Fi framework that leverages multi-LLM-agent systems to address the complex, dynamic challenges of MAPC. By empowering each AP with a cognitive architecture featuring reasoning, memory, and tool use, we demonstrated that agents can autonomously learn to navigate the trade-off between Co-TDMA and Co-SR through natural language dialogue. The primary contribution of this work is a new paradigm for decentralized wireless control that transcends the limitations of both rigid protocols and small-model AI. Our comprehensive simulations validate that this approach not only outperforms optimized conventional schemes but also exhibits strong adaptability and backward compatibility. Future work will focus on endowing these agents with capabilities for continuous self-improvement and strategy evolution, further paving the way for truly intelligent and self-organizing wireless networks.

\bibliographystyle{IEEEtran}
\bibliography{IEEEabrv,reference}

\begin{thebibliography}{10}
\providecommand{\url}[1]{#1}
\csname url@samestyle\endcsname
\providecommand{\newblock}{\relax}
\providecommand{\bibinfo}[2]{#2}
\providecommand{\BIBentrySTDinterwordspacing}{\spaceskip=0pt\relax}
\providecommand{\BIBentryALTinterwordstretchfactor}{4}
\providecommand{\BIBentryALTinterwordspacing}{\spaceskip=\fontdimen2\font plus
\BIBentryALTinterwordstretchfactor\fontdimen3\font minus \fontdimen4\font\relax}
\providecommand{\BIBforeignlanguage}[2]{{%
\expandafter\ifx\csname l@#1\endcsname\relax
\typeout{** WARNING: IEEEtran.bst: No hyphenation pattern has been}%
\typeout{** loaded for the language `#1'. Using the pattern for}%
\typeout{** the default language instead.}%
\else
\language=\csname l@#1\endcsname
\fi
#2}}
\providecommand{\BIBdecl}{\relax}
\BIBdecl

\bibitem{galati2024primer}
L.~Galati-Giordano, G.~Geraci, M.~Carrascosa, and B.~Bellalta, ``What will {Wi-Fi} 8 be? a primer on {IEEE} 802.11bn ultra high reliability,'' \emph{IEEE Commun. Mag.}, vol.~62, no.~8, pp. 126--132, Aug. 2024.

\bibitem{tgbn}
R.~J. Yu, ``{Specification Framework for TGbn},'' Sep. 2024, [Online]. Available: \url{https://mentor.ieee.org/802.11/dcn/24/11-24-0209-05-00bn-specification-framework-for-tgbn.docx}.

\bibitem{cosr}
{IEEE 802.11 22/1822r0}, ``Recap on coordinated spatial reuse operation,'' IEEE 802.11 Working Group document, May 2022, [Online]. Available: \url{https://mentor.ieee.org/802.11/dcn/22/11-22-1822-00-0uhr-overhead-analysis-of-coordinated-spatial-reuse.pptx}.

\bibitem{Wilhelmi2023thr}
F.~Wilhelmi, L.~Galati-Giordano, G.~Geraci, B.~Bellalta, G.~Fontanesi, and D.~Nuñez, ``Throughput analysis of {IEEE} 802.11bn coordinated spatial reuse,'' in \emph{Proc. IEEE CSCN}, Nov. 2023, pp. 401--407.

\bibitem{wojnar2024ieee}
M.~Wojnar \emph{et~al.}, ``{IEEE} 802.11 bn multi-{AP} coordinated spatial reuse with hierarchical multi-armed bandits,'' \emph{IEEE Commun. Lett.}, Dec. 2024.

\bibitem{yu2025hmarl}
J.~Yu, L.~Liang, H.~Ye, and S.~Jin, ``Hierarchical multi-agent reinforcement learning-based coordinated spatial reuse for next generation {WLANs},'' \emph{arXiv preprint arXiv:2506.14187}, 2025.

\bibitem{liang2025wirelessllm}
L.~Liang \emph{et~al.}, ``Large language models for wireless communications: From adaptation to autonomy,'' \emph{arXiv preprint arXiv:2507.21524}, 2025.

\bibitem{tong2025wirelessagent}
J.~Tong \emph{et~al.}, ``{WirelessAgent}: Large language model agents for intelligent wireless networks,'' \emph{arXiv preprint arXiv:2505.01074}, 2025.

\bibitem{icl}
S.~Min, X.~Lyu, A.~Holtzman, M.~Artetxe, M.~Lewis, H.~Hajishirzi, and L.~Zettlemoyer, ``Rethinking the role of demonstrations: What makes in-context learning work?'' in \emph{Proc. EMNLP}, Dec. 2022.

\bibitem{cot}
J.~Wei, X.~Wang, D.~Schuurmans, M.~Bosma, B.~Ichter, F.~Xia, E.~Chi, Q.~Le, and D.~Zhou, ``Chain of thought prompting elicits reasoning in large language models,'' in \emph{Proc. NeurIPS}, Dec. 2022.

\bibitem{residential}
S.~Merlin \emph{et~al.}, ``{TGax Simulation Scenarios},'' Jul. 2015, [Online]. Available: \url{https://mentor.ieee.org/802.11/dcn/14/11-14-0980-16-00ax-simulation-scenarios.docx}.

\bibitem{openai2024gpt4ocard}
OpenAI \emph{et~al.}, ``{GPT}-4o system card,'' \emph{arXiv preprint arXiv:2410.21276}, 2024.

\bibitem{deepseekai2025deepseekr1}
DeepSeek-AI \emph{et~al.}, ``{DeepSeek-R1}: Incentivizing reasoning capability in {LLMs} via reinforcement learning,'' \emph{arXiv preprint arXiv:2501.12948}, 2025.

\bibitem{wu2023autogen}
Q.~Wu \emph{et~al.}, ``{AutoGen}: Enabling next-gen {LLM} applications via multi-agent conversation,'' \emph{arXiv preprint arXiv:2308.08155}, 2023.

\end{thebibliography}

\end{document}